%% file: main.tex
\documentclass{article}

\usepackage{microtype}
\usepackage{graphicx}
\usepackage{subfigure}
\usepackage{booktabs}
\usepackage[usenames,dvipsnames]{xcolor}
\usepackage[normalem]{ulem}

\usepackage{algorithm,algorithmic}

\usepackage[colorlinks=true,citecolor=blue]{hyperref}

\usepackage[left=1.25in, top=1in, bottom=1in, right=1.25in]{geometry}
\input{macros.tex}

\title{Doubly-stochastic mining for heterogeneous retrieval}
\author{Ankit Singh Rawat \and Aditya Krishna Menon \and Andreas Veit \and Felix Yu \and Sashank J. Reddi \qquad Sanjiv Kumar \\
Google Research, New York \\
\texttt{\{ankitsrawat,adityakmenon,aveit,felixyu,sashank,sanjivk\}@google.com}}

\begin{document}

\maketitle

\input{body}

\end{document}

%% file: macros.tex
\usepackage[utf8]{inputenc} 
\usepackage[T1]{fontenc}    
\usepackage[colorlinks=true,citecolor=blue]{hyperref}       
\usepackage{url}            
\usepackage{booktabs}       
\usepackage{amsfonts}       
\usepackage{nicefrac}       
\usepackage{microtype}      

\usepackage{natbib}

\usepackage{tikz,pgfplots}
\usetikzlibrary{fadings}

\usepackage{float,subfloat}
\usepackage[font=footnotesize,format=hang]{caption}
\usepackage{graphicx}
\usepackage[export]{adjustbox}
\usepackage[inline]{enumitem}

\usepackage{tikz}
\usetikzlibrary{positioning}

\usepackage{amsmath,amssymb,amsthm}
\usepackage[mathscr]{eucal}
\usepackage{stmaryrd}
\usepackage{accents}
\usepackage{upgreek}

\usepackage[most]{tcolorbox}
\usepackage{setspace}

\usepackage{booktabs,colortbl,multirow}

\graphicspath{{figs//}}

\allowdisplaybreaks

\newcommand\numberthis{\addtocounter{equation}{1}\tag{\theequation}}

\newcommand\todo[1]{{\scriptsize\textcolor{ForestGreen}{\textbf{TODO:} #1}}}

\newcommand{\defEq}{\stackrel{.}{=}}

\newcommand{\indicator}[1]{\llbracket #1 \rrbracket}

\newcommand{\tick}{$\checkmark$}
\newcommand{\cross}{$\times$}

\newcommand{\SSNM}{S$^2$M}
\newcommand{\best}[1]{\cellcolor{gray!25}{#1}}

\newcommand{\ellSNM}{\ell_{\mathrm{snm}}}

\newcommand{\Minibatch}{S_{\mathrm{mb}}}
\newcommand{\MinibatchSize}{N_{\mathrm{mb}}}

\newcommand{\avgTopKPrime}{\widehat{L}_{\mathrm{top}( k' )}}
\newcommand{\pAvgTopKPrime}{{L}_{\mathrm{top}( k' )}}

\newcommand{\kExample}{k'}
\newcommand{\kLabel}{k}

\newcommand{\NumLabels}{K}

\newcommand{\argmin}{{\operatorname{argmin }}}

\renewcommand{\Pr}{\mathbb{P}}
\newcommand{\E}[2]{\underset{#1}{\mathbb{E}}\left[ #2 \right]}

\newcommand{\lenet}{{\sc LeNet}}
\newcommand{\resnet}{{\sc ResNet}}
\newcommand{\mnist}{{\sc MNIST}}
\newcommand{\cifarsmall}{{\sc CIFAR-10}}
\newcommand{\amazonsmall}{{\sc AmazonCat-13k}}

\newcommand{\wikilshtc}{{\sc WikiLSHTC}}

\newcommand{\X}{\mathsf{X}}

\newcommand{\Y}{\mathsf{Y}}

\newcommand{\DCal}{\mathscr{D}}

\newcommand{\FCal}{\mathscr{F}}

\newcommand{\OCal}{\mathscr{O}}

\newcommand{\SCal}{\mathscr{S}}
\newcommand{\TCal}{\mathscr{T}}
\newcommand{\UCal}{\mathscr{U}}

\newcommand{\XCal}{\mathscr{X}}
\newcommand{\YCal}{\mathscr{Y}}

\newcommand{\Real}{\mathbb{R}}

\newcommand{\cvar}{\mathrm{CVaR}}

\newcommand{\avgLoss}{\widehat{L}_{\mathrm{avg}}}
\newcommand{\avgTopK}{\widehat{L}_{\mathrm{top}(k)}}

\newcommand{\avgTop}[2]{\widehat{L}_{{\mathrm{top}(#1)}}( f; #2 )}

\newcommand{\XIYI}{( x_i, y_i )}

\newtheorem{lemma}{Lemma}

\newtheorem{proposition}[lemma]{Proposition}

\theoremstyle{definition}

\newtcbtheorem[number within=section]%
{remark_tcb} 
{Remark} 
{%
fonttitle=\bfseries\upshape,fontupper=\normalsize,
colframe=green!10!black,colback=green!2!white,
colbacktitle=green!20!white,coltitle=blue!75!black,
boxsep=1pt,left=2pt,right=2pt,top=1pt,bottom=1pt,
frame hidden,boxrule=1pt,
theorem style=plain} 
{rem} 

\newtcbtheorem[number within=section]%
{assumption_tcb} 
{Assumption} 
{%
fonttitle=\bfseries\upshape,fontupper=\normalsize,
colframe=blue!5!black,colback=blue!2!white,
colbacktitle=blue!10!white,coltitle=blue!75!black,
boxsep=1pt,left=2pt,right=2pt,top=1pt,bottom=1pt,
frame hidden,boxrule=1pt,
theorem style=plain} 
{asm} 

\newtcbtheorem[number within=section]%
{example_tcb} 
{Example} 
{%
fonttitle=\bfseries\upshape,fontupper=\normalsize,
colframe=green!10!black,colback=green!2!white,
colbacktitle=green!20!white,coltitle=blue!75!black,
boxsep=1pt,left=2pt,right=2pt,top=1pt,bottom=1pt,
frame hidden,boxrule=1pt,
theorem style=plain} 
{ex} 

%% file: body.tex
\begin{abstract}
\input{abstract}
\end{abstract}

\section{Introduction}
\input{intro}


\section{Background and notation}
\label{sec:background}
\input{formulation}
\section{Doubly-stochastic mining (S$^2$M)}
\label{sec:algorithm}
\input{method}

\section{Analysis of S$^2$M under heterogeneity}
\label{sec:analysis}
\input{analysis}

\section{Experiments}
\label{sec:experiments}
\input{experiments}

\section{Conclusion and future work}
\input{conclusion.tex}

\clearpage

%

\bibliography{references}
\bibliographystyle{plainnat}

\clearpage
\appendix
\onecolumn

\section{Proofs}
\label{appen:proofs}
\input{proofs}

%% file: abstract.tex
Modern retrieval problems are characterised by training sets with potentially billions of labels, and heterogeneous data distributions across \emph{subpopulations} (e.g., users of a retrieval system may be from different countries), each of which poses a challenge. 
The first challenge concerns \emph{scalability}: 
with {{a large number}} of labels, standard losses are difficult to optimise even on a single example.
The second challenge concerns \emph{uniformity}: one ideally wants good performance on each subpopulation. 
While several solutions have been proposed to address the first challenge, the second challenge has received relatively less attention.
In this paper, we propose \emph{doubly-stochastic mining} (\emph{\SSNM{}}), a stochastic optimization technique that addresses both challenges. 
In each iteration of \SSNM{}, 
we compute a per-example loss based on a subset of hardest \emph{labels},
and then compute the minibatch loss based on the hardest \emph{examples}. 
We show theoretically and empirically that 
by focusing on the hardest examples, \SSNM{} ensures that 
\emph{all}
data subpopulations are modelled well.

%% file: intro.tex
Information retrieval concerns finding documents 
that are most relevant for a given query,
and is a canonical real-world use case for machine learning~\citep{Manning:2008}.
The simplest incarnation of retrieval models involves learning a real-valued scoring function that ranks, 
for each example, 
the set of possible labels it may be matched to.
A core challenge is \emph{scalability}:
there may be billions of 
examples (e.g., user queries) and
labels (e.g., videos in a recommendation system),
each of whose scores na\"{i}vely needs to be updated at every training iteration.
Effective means of addressing both problems have been widely studied~\citep{Mikolov:2013,Jean:2015,Reddi:2019}.

A distinct challenge is \emph{heterogeneity}:
the distribution over examples is often a mixture of diverse \emph{subpopulations} (e.g., queries may arise from geographically disparate user bases).
Na\"{i}ve training on such data may lead 
to models that
perform disproportionately well on one subpopulation at the expense of others;
e.g., if queries originate from multiple countries,
the retrieval model may only perform well on queries from the dominant country.
Such behaviour is clearly undesirable.

Unfortunately,
the heterogeneity problem plagues (to our knowledge) \emph{all} state-of-the-art retrieval models.
Indeed,
while there has been considerable progress on improving the efficacy of retrieval systems~\citep{Prabhu:2018,Reddi:2019,Guo:2019},
\emph{any} such method reliant on optimising examples drawn uniformly at random from the training set (e.g., via SGD)
will inherently be biased towards the ``dominant'' subpopulations. This raises the natural question:
\emph{Can one mitigate such bias
without introducing significant computational overhead?}

\begin{figure*}[!t]
    \centering
    
    {
        \includegraphics[scale=0.75,valign=b]{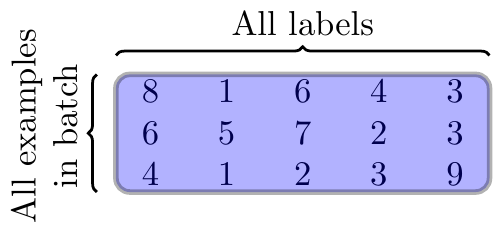}
    }
    {
        \includegraphics[scale=0.75,valign=b]{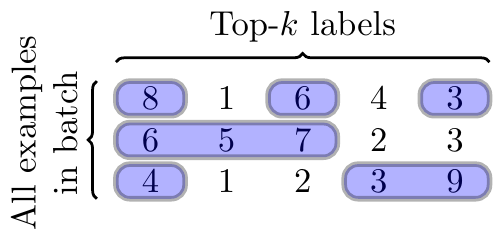}
    }
    {
        \includegraphics[scale=0.75,valign=b]{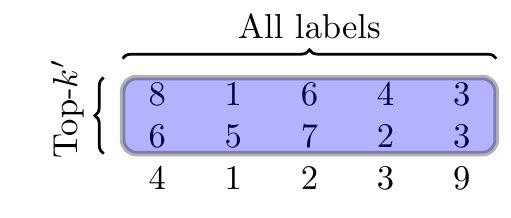}
    }
    {
        \includegraphics[scale=0.75,valign=b]{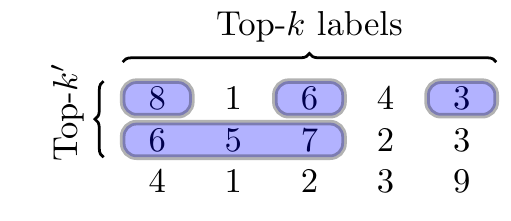}
    }    
    
    \caption{
    Illustration of
    stochastic gradient descent (SGD),
    stochastic negative mining (SNM),
    average top-$\kExample$ SGD ($q$-SGD),
    and our proposed doubly stochastic mining (\SSNM{}).
    The matrices depict losses computed over a minibatch of three instances and five labels,
    with the per-example loss averaging over the loss for each label.
    In SGD, we compute the average loss over each instance, which in turn averages over each of the labels.
    In SNM, we only retain labels with 
    the $\kLabel$-highest contributions to the per-example losses;
    here, $\kLabel = 3$.
    In top-$\kExample$ SGD, we only retain examples with top-$\kExample$ highest loss; here $\kExample = 2$.
    In \SSNM{}, one combines both of these, allowing tractable training under a large number of labels,
    as well as good performance on subpopulations.
    }
    \label{fig:summary}
\end{figure*}

In this paper, we affirmatively answer this question,
and propose a simple algorithm which
ensures good performance across different subpopulations,
while
scaling to large-scale retrieval settings.
Specifically, our contributions are:
\begin{enumerate}[label=(\roman*),itemsep=0pt,topsep=0pt]
    \item we propose \emph{doubly-stochastic mining} (\SSNM{}),
    an algorithm that
    optimises a loss computed over the \emph{hardest} examples {and} labels
    in a randomly drawn minibatch.

    \item we establish that \SSNM{} controls the retrieval loss over latent subpopulations,
    drawing on
    a connection between our objective
    and
    the \emph{conditional value-at-risk} (\emph{CVaR}).

    \item we show that empirically, \SSNM{} 
    yields improved retrieval performance on subpopulations compared to state-of-the-art retrieval methods.
\end{enumerate}

In more detail,
given a minibatch of examples,
\SSNM{}
randomly draws a sample of \emph{labels},
from which we 
compute a per-example loss based on the hardest labels in this subset;
we
then average the hardest per-example losses to form the minibatch loss.
Algorithmically,
this is a combination of two distinct proposals in the literature (see Figure~\ref{fig:summary}):
the \emph{stochastic negative mining} (\emph{SNM}) algorithm of~\citep{Reddi:2019},
and \emph{$q$-SGD}~\citep{Kawaguchi:2019},
which focus on the hardest labels and examples, respectively.

While combining these two algorithms is conceptually straightforward,
it is far less obvious that the result can cope with the heterogeneity problem.
Contribution (ii) 
exploits two non-apparent links 
---
that of the \SSNM{} objective to the conditional value-at-risk (CVaR)~\citep{Rockafellar:2000},
and of the CVaR to a worst-case loss over subpopulations
---
to
establish that by focusing on the hardest examples, 
\SSNM{} ensures good performance on latent data subpopulations.
Our analysis integrates recent results from 
both the fairness~\citep{Mohri:2019}
and the operations research~\citep{Cherukuri:2019} literature.

The idea of focusing on the hardest examples 
has more broadly been explored in several different contexts.
For example,
minimising the maximal loss forms the basis for the hard-margin SVM~\citep{Vapnik:1999},
and has more broadly been studied in~\citet{Shalev-Shwartz:2016}.
However, these works do not consider jointly sampling instances and labels,
and are thus not attuned to the retrieval setting of interest.
Active learning also concerns finding ``hard'' examples to label~\citep{Dagan:1995},
but in an interactive setting distinct to 
our passive retrieval setting.

%% file: formulation.tex
We review the multiclass retrieval setting of interest\footnote{Our results easily extend to multilabel setting via suitable reductions~(see, e.g.,~\citet{Wydmuch:2018}).},
and approaches to address the underlying challenges.
Table~\ref{tbl:glossary} summarises some commonly used symbols.

\subsection{Multiclass retrieval}

In multiclass classification,
we observe instances $x \in \XCal$ with associated labels $y \in \YCal = [ \NumLabels ] \defEq \{ 1, 2, \ldots, \NumLabels \}$
drawn from some distribution $\DCal$ over $\XCal \times \YCal$.
We aim to learn a \emph{scorer} $f \colon \XCal \to \Real^\NumLabels$ which can order the labels by their relevance for an instance.
In a retrieval setting, our goal is to 
minimise the \emph{retrieval risk}, 
which for an integer $r$ is
\begin{equation}
    \label{eqn:top-k-risk}
    L_{\mathrm{ret}(r)}( f; \DCal ) 
    \defEq \E{( \X, \Y ) \sim \DCal}{ \ell_{\mathrm{ret}(r)}( \Y, f( \X ) ) },
\end{equation}
where $\ell_{\mathrm{ret}(r)}( i, f ) \defEq \indicator{i \notin \mathrm{top}_r( f )}$, 
and $\mathrm{top}_r( f )$ denotes the $r$ indices with highest score under $f$.
When $r = 1$, this reduces to the misclassification risk $L_{01}( f; \DCal )$.

%

In practice, directly minimising $L_{\mathrm{ret}(r)}( f; \DCal )$ is computationally prohibitive
owing to the non-differentiability of the loss $\ell_{\mathrm{ret}( r )}$.
This can be alleviated by minimising a suitable \emph{surrogate} loss $\ell \colon [ \NumLabels ] \times \Real^\NumLabels \to \Real_+$.
Examples of $\ell$ include the softmax cross-entropy 
$\ell( y, f( x ) ) = -f_y( x ) + \log \sum_{y'} e^{f_{y'}( x )}$,
and 
\emph{binary ordered weighted losses} (\emph{BOWLs})~\citep{Reddi:2019}
\begin{equation}
    \label{eqn:bowl}
    \ell( y, f( x ) ) = \phi( f_y( x ) ) + \mathrm{avg}_{\mathrm{top}( k )}\left( \{ \phi( -f_{y'}( x ) ) \}_{y' \neq y} \right),
\end{equation}
where 
$\phi \colon \Real \to \Real_+$ is a margin loss (e.g., hinge $\phi( z ) = [ 1 - z ]_+$),
$k$ is an integer,
and $\mathrm{avg}_{\mathrm{top}( k )}$ denotes the average of the top-$k$ largest elements of a vector.
When $k = 1$,
this is the Cramer-Singer loss~\citep{Crammer:2002},
\begin{equation}
    \label{eqn:cs-loss}
    \ell( y, f( x ) ) = \phi( f_y( x ) ) + \max_{y' \neq y} \phi( -f_{y'}( x ) ).
\end{equation}
This provides a tight bound on $L_{\mathrm{ret}(1)}( f; \DCal )$.
When $k = \NumLabels - 1$,
this is the averaged loss~\citep{Zhang:2004b},
\begin{equation}
    \label{eqn:ova}
    \ell( y, f( x ) ) = \phi( f_y( x ) ) + \frac{1}{\NumLabels - 1} \sum_{j \neq y} \phi( -f_j( x ) ).
\end{equation}
For a training set $S \defEq \{ \XIYI \}_{i \in [N]} \sim \DCal^N$
and surrogate loss $\ell$,
one may seek to minimise the empirical risk
\begin{equation}
    \label{eqn:empirical-risk}
    \avgLoss( f; S ) \defEq \frac{1}{N} \sum_{i \in [N]} \ell( y_i, f( x_i ) ).
\end{equation}

\begin{table}[!t]
    \centering
    \renewcommand{\arraystretch}{1.25}
    \small
    \begin{tabular}{@{}lp{2in}@{}}
        \toprule
        \textbf{Symbol} & \textbf{Meaning} \\
        \toprule
        $N, \NumLabels, \bar{\NumLabels}$ & \# of examples, labels, sampled labels \\
        $\kExample, \kLabel$ & \# of examples and labels used for top-averaging \\
        $\ell, \ellSNM$ & Generic and SNM multiclass loss \\
        $S, \Minibatch$ & Training set and minibatch \\
        $L_{\mathrm{top}( \kExample )}, \widehat{L}_{\mathrm{top}( \kExample )}$ & Population and empirical top-$\kExample$ loss \\
        \bottomrule
    \end{tabular}
    \caption{Glossary of commonly used symbols.}
    \label{tbl:glossary}
\end{table}

\subsection{Challenges in multiclass retrieval}
\label{sec:challenges}

Multiclass retrieval poses several challenges.
First, the number of samples $N$ can potentially be in the order of billions, thus making computing $\avgLoss( f; S )$ prohibitive~\citep{Bottou:2007}.
Second, the number of labels $\NumLabels$ may also be in the order of billions, thus making even computing a \emph{single} $\ell( y_i, f( x_i ) )$ prohibitive~\citep{Agrawal:2013,Yu:2014,Bhatia:2015,Jain:2016,Babbar:2017,Prabhu:2018,Jain:2019};
e.g., the softmax cross-entropy requires computing $\log \sum_{y' \in [L]} e^{f_{y'}( x_i )}$, 
which is expensive for large $\NumLabels$.

Third, the distribution $\DCal$ typically comprises heterogeneous subpopulations;
e.g., 
the users of the retrieval system may be from different countries.
Formally, suppose
$\DCal = \sum_{p \in [P]}\nu_{p} \cdot \DCal_p$
for $P$
distributions $\{\DCal_p\}_{p \in [P]}$
with
mixture weights $\nu \in \Delta_P$,
where $\Delta_P$ denotes the simplex.
Our goal is to ensure good performance for \emph{each} $\DCal_p$.

We now review proposals to deal with each challenge.

\subsection{Stochastic gradient descent and negative mining}

When $N$ is large,
one may perform 
\emph{minibatch stochastic gradient descent} (\emph{SGD})~\citep{Robbins:1951}
by sampling
a \emph{minibatch} $\Minibatch = \{ ( x_j, y_j ) \}_{j = 1}^{\MinibatchSize} \subseteq S$, 
and performing descent based on the \emph{average minibatch loss}:
\begin{equation}
    \label{eqn:minibatch-loss}
    \avgLoss( f; \Minibatch ) \defEq \frac{1}{\MinibatchSize} \sum_{j \in [ \MinibatchSize ]} \ell( y_j, f( x_j ) ).
\end{equation}

%
\label{sec:snm}

When $\NumLabels$ is large,
however,
it is challenging to use SGD since even \emph{computing} the minibatch loss may be prohibitive, let alone optimising it.
As noted above,
to compute the loss on even a \emph{single} example,
the softmax cross-entropy
as well as
the BOWL losses in~\eqref{eqn:bowl}
require the scores of all $\NumLabels$ labels,
which may not be feasible.

One approach to cope with this problem is to \emph{sample} the labels, and use a resulting stochastic approximation to the loss~\citep{Bengio:2008,Jean:2015,Grave:2017}.
In \emph{stochastic negative mining} (SNM)~\citep{Reddi:2019},
in addition to the minibatch of random examples, 
for each example
one draws
a {random subset of} \emph{``negative'' labels} $\bar{\YCal} \subseteq \YCal - \{ y \}$ of size $\bar{\NumLabels} \ll \NumLabels$.
One can then treat $\bar{\YCal}$ as the entire label space,
and compute a new BOWL 
\begin{equation}
    \label{eqn:snm-bowl}
    \begin{aligned}
        \ellSNM( y, f( x ); \bar{\YCal} ) = \phi( f_y( x ) ) + 
        \mathrm{avg}_{\mathrm{top}( \kLabel )}( \{ \phi( -f_{y'}( x ) ) \}_{y' \in \bar{\YCal}} ).
    \end{aligned}
\end{equation}
As a simple example, 
for $\kLabel = 1$,
\begin{equation}
    \label{eqn:snm-max}
    \ellSNM( y, f( x ); \bar{\YCal} ) = \phi( f_y( x ) ) + \max_{y' \in \bar{\YCal}} \phi( -f_{y'}( x ) ),
\end{equation}
where, by contrast to~\eqref{eqn:cs-loss}, the maximum is only over the labels in $\bar{\YCal}$.
A key property of SNM is that one only needs to update $O(\bar{\NumLabels})$ labels' parameters,
which can be a significant saving compared to updating the parameters for all $\NumLabels$ labels.

Observe that $\ellSNM$ is stochastic, owing to $\bar{\YCal}$ being random.
The expected loss
is itself a BOWL, 
which 
is
calibrated~\citep{Zhang:2004b} under mild conditions.

%
\subsection{Maximum empirical loss}
\label{sec:avg-top-k-loss}

Both SGD and SNM take 
minimising~\eqref{eqn:empirical-risk} to be their basic goal.
This
involves minimising the average loss over the training set. 
One may however replace the average with other summaries,
such as
the \emph{maximum}:
\begin{equation}
    \label{eqn:max-loss}
    \widehat{L}_{\mathrm{max}}( f; S ) \defEq \max_{i \in [N]} \ell( y_i, f( x_i ) ).
\end{equation}

\citet{Shalev-Shwartz:2016} established that 
the minimiser of~\eqref{eqn:max-loss} guarantees good performance on subpopulations.
Intuitively,~\eqref{eqn:max-loss} encourages predicting well on \emph{all} examples,
including those from a rare subpopulation.
The authors also proposed a means of stochastically optimising the objective,
but this requires $\mathscr{O}( N )$ auxiliary variables,
which can be prohibitive in large-scale settings.

%% file: method.tex
We now present \SSNM{}, 
our doubly-stochastic mining algorithm,
which addresses all three challenges discussed above:
\SSNM{} scales to settings with a large number of examples, labels, and 
performs well on heterogeneous subpopulations.

\subsection{Doubly-stochastic mining}

Reviewing existing multiclass retrieval methods,
we see that they fail to meet one or more of the challenges in~\S\ref{sec:challenges} (see Table~\ref{tbl:summary} for a summary);
e.g., while SNM 
handles large $N$ and $\NumLabels$, it
does not adapt to heterogeneous distributions.

\begin{table*}[!t]
    \centering
    \renewcommand{\arraystretch}{1.25}
    
    \resizebox{0.99\linewidth}{!}{
    \begin{tabular}{@{}lllll}
        \toprule
        \textbf{Method} & \textbf{Loss on minibatch} & \textbf{Large $N$?} & \textbf{Large $\NumLabels$?} & \textbf{Heterogenity?} 
        \\
        \toprule
        SGD~\citep{Robbins:1951} & $\displaystyle \mathrm{avg}\left( \left\{ \phi(  f_{y}( x ) ) + \mathrm{avg}\left( \{ \phi( -f_{y'} ) \}_{y' \in \YCal} \right) \right\} \right)$ & \tick & \cross & \cross 
        \\
        SNM~\citep{Reddi:2019} & $\displaystyle \mathrm{avg}\left( \phi(  f_{y}( x ) ) + \mathrm{avg}_{\mathrm{top}( k )}\left( \{ \phi( -f_{y'} ) \}_{y' \in \bar{\YCal}} \right) \right)$ & \tick & \tick & \cross 
        \\
        Top-$\kExample$ SGD~\citep{Kawaguchi:2019} & $\displaystyle \mathrm{avg}_{\mathrm{top}( \kExample )}\left( \left\{ \phi(  f_{y}( x ) ) + \mathrm{avg}\left( \{ \phi( -f_{y'} ) \}_{y' \in \YCal} \right) \right\} \right)$ & \tick & \cross & \tick 
        \\
        \SSNM{}~(This paper) & $\displaystyle \mathrm{avg}_{\mathrm{top}( \kExample )}\left( \left\{ \phi(  f_{y}( x ) ) + \mathrm{avg}_{\mathrm{top}( k )}\left( \{ \phi( -f_{y'} ) \}_{y' \in \bar{\YCal}} \right) \right\} \right)$ & \tick & \tick & \tick 
        \\
        \bottomrule
    \end{tabular}
    }
    \caption{Comparison of losses employed in 
    stochastic gradient descent (SGD),
    stochastic negative mining (SNM),
    average top-$\kExample$ SGD (top-$\kExample$ SGD),
    and our proposed doubly stochastic mining (\SSNM{}).
    Here, $\Minibatch$ denotes a random minibatch of (instance, label) pairs $(x, y)$,
    and $f( x ) \in \Real^\NumLabels$ the model predictions for each of $\NumLabels$ possible classes.
    We assume the use of a base convex loss $\phi \colon \Real \to \Real_+$
    used to construct a multiclass loss per~\eqref{eqn:bowl}.
    In SGD, one computes the average loss over the minibatch, which implicitly averages over all labels.
    In SNM, one 
    draws a random batch of ``negative'' labels $\bar{\YCal} \subseteq \YCal - \{ y \}$,
    and 
    only operates on the top-$\kLabel$ highest scoring labels,
    denoted by the $\mathrm{avg}_{\mathrm{top}(k)}( \cdot )$ operation.
    In top-$\kExample$ SGD, one only operates on the top-$\kExample$ highest per-example losses,
    but for each per-example loss averages over all labels.
    In \SSNM{}, one combines both of these, with the aim of allowing tractable training under a large $\NumLabels$,
    while ensuring good performance on heterogeneous data subpopulations.}
    \label{tbl:summary}
\end{table*}

A natural question is whether we can extend SNM to rectify this.
Following~\S\ref{sec:avg-top-k-loss},
we take inspiration from~\citet{Shalev-Shwartz:2016}
and move from \emph{average} to the \emph{maximal} empirical loss.
Indeed, 
given 
a minibatch $\Minibatch = \{ ( x_j, y_j ) \}_{j = 1}^{\MinibatchSize}$,
the analogue of~\eqref{eqn:max-loss} is:
\begin{equation}
    \label{eqn:minibatch-max}
    \widehat{L}_{\mathrm{max}}( f; \Minibatch ) \defEq \max_{j \in [ \MinibatchSize ]} \ell( y_j, f( x_j ) ).
\end{equation}
Interestingly, 
this can be seen as an example level counterpart to the SNM loss of~\eqref{eqn:snm-max}:
in the latter, we compute a {per-example loss} $\ellSNM( y, f( x ) )$ via the maximal loss over all \emph{labels} in $\bar{\YCal}$,
while in~\eqref{eqn:minibatch-max}, 
{we compute the minibatch loss $\widehat{L}_{\mathrm{max}}( f; \Minibatch )$ via 
the maximal loss
over all \emph{examples} 
in the minibatch.} 
To further highlight this,
suppose we replace the maximum with the average of the \emph{$\kExample$ largest losses}:
\begin{equation}
    \label{eqn:minibatch-top-k}
    \widehat{L}_{\mathrm{top}( \kExample )}(f; \Minibatch) \defEq \mathrm{avg}_{\mathrm{top}( \kExample )}( \{ \ell( y_j, f( x_j ); \bar{\YCal} ) \}_{j \in [ \MinibatchSize ]} ).
\end{equation}
This is in contrast to~\eqref{eqn:snm-bowl}, 
{which obtains a per-example loss by 
averaging over the $\kLabel$ labels with the highest 
contribution to the loss value.} 
By combining the SNM loss of~\eqref{eqn:snm-bowl} with the top-$\kExample$ minibatch loss of~\eqref{eqn:minibatch-top-k}, we have a \emph{doubly stochastic} procedure which handles both large $\NumLabels$ and heterogeneous example distributions.
Intuitively, this should behave similarly to
the maximum loss in~\eqref{eqn:minibatch-max} in controlling subpopulation performance 
while being more noise-robust.

Formally,
our \emph{doubly-stochastic mining} (\SSNM{}) algorithm\footnote{The name signifies that given a minibatch, 
the algorithm mines both the hardest labels as well as examples.} 
with a training set 
$S \sim \DCal^N$
is summarised in Algorithm~\ref{alg:snm2}.
In a nutshell, the procedure is as follows:
first, following SGD, we draw a random minibatch 
$\Minibatch \subseteq S$.
{Next, following SNM, 
for a given example $(x, y) \in \Minibatch$, we draw a random sample of $\bar{\NumLabels} \ll \NumLabels$ labels 
$\bar{\YCal} \subseteq \YCal - \{ y \}$, and compute per-example loss $\ellSNM( y, f( x ))$, per~\eqref{eqn:snm-bowl};
this will only focus on the hardest $\kLabel$ labels within $\bar{\NumLabels}$. 
Given $\{ \ellSNM( y, f( x ); \bar{\YCal} ) \colon ( x, y ) \in \Minibatch \}$, we average the top-$\kExample$ loss values, and use this as our minibatch loss, per~\eqref{eqn:minibatch-top-k}.} 

We emphasise that \SSNM{} involves two forms of top-$k$ averaging, with different purposes.
In constructing the SNM loss $\ellSNM$, we only average over the top-$\kLabel$ highest scoring labels.
This is motivated by yielding both computational tractability,
and a tight bound to the 
retrieval loss.
In constructing the final minibatch loss, we only average over the top-$\kExample$ highest per-sample losses.
This is motivated by controlling the loss over {subpopulations};
we formalise this in~\S\ref{sec:analysis}.

Further, $\bar{\NumLabels}$ and $\kLabel$ may be picked following~\citet{Reddi:2019}: 
to ensure good retrieval performance, one can pick $\bar{\NumLabels}$ based on computational considerations, and $\kLabel = r$ for the retrieval threshold $r$ in~\eqref{eqn:top-k-risk}.
The choice of $\kExample$ is more subtle:
{the discussion following
Proposition~\ref{prop:cvar-gen-bound} in \S\ref{sec:analysis}} reveals that $\kExample$ should correspond to the number of samples from the rarest subpopulation one expects to see in $S$.
In practice, this {can be} specified as a tuning parameter.

\subsection{Special cases of \SSNM{}}
\label{sec:special}

\SSNM{} generalises several existing approaches:
\begin{enumerate}[label=(\alph*),itemsep=0pt,topsep=0pt]
    \item when $\bar{\NumLabels} = \NumLabels$ and $\kExample = \MinibatchSize$, we recover SGD.
    
    \item when $\bar{\NumLabels} \ll \NumLabels$ and $\kExample = \MinibatchSize$, we recover SNM.
    
    \item when $\bar{\NumLabels} = \NumLabels$ and $\kExample \ll \MinibatchSize$, we recover average top-$\kExample$ SGD~\citep{Kawaguchi:2019}.
\end{enumerate}
Our interest is in 
focussing on
both the hardest per-example losses ($\kExample \ll \MinibatchSize$),
\emph{and} 
the hardest labels within each such loss ($k \ll \bar{\NumLabels} \ll \NumLabels$).
For point (c) above,
for the small $\NumLabels$ setting,~\citet{Fan:2017,Kawaguchi:2019} considered minimising the average top-$\kExample$ loss per~\eqref{eqn:minibatch-top-k},
with the latter specifically focussed on the stochastic setting.
However, these works did \emph{not} consider sampling over labels,
nor theoretically justify the ability of top-$\kExample$ losses to handle heterogeneous example distributions (as we shall do in~\S\ref{sec:analysis}).

Compared to SNM, \SSNM{} replaces the average of \emph{all} per-example losses with the average of the \emph{top-$\kExample$} per-example losses. 
For a minibatch size of $\MinibatchSize$, this step becomes $\mathscr{O}( \MinibatchSize \cdot \log \MinibatchSize )$ rather than $\mathscr{O}( \MinibatchSize )$.
Since $\MinibatchSize \ll N$ is typically a small constant, the overhead is minimal.

%

\begin{algorithm}[!t]
    \caption{Doubly-stochastic mining (\SSNM{})}
    \label{alg:snm2}
    \small
    \begin{algorithmic}[1]
        \STATE \textbf{Input}: $S \defEq \{ ( x_i, y_i ) \}_{i = 1}^N$, training steps $T$, minibatch size $\MinibatchSize$, label sample size $\bar{\NumLabels}$, $k, \kExample \in \mathbb{N}_+$

        \FOR{$t = 1, 2, \ldots, T$}
            \STATE draw minibatch $\Minibatch = \{ ( x_j, y_j ) \}_{j = 1}^{\MinibatchSize} \subseteq S$
            \STATE draw label sample $\bar{\YCal}_j \subseteq \YCal - \{ y_j \}$ of size $\bar{\NumLabels}$
            $\forall j$
            \STATE compute $\ell_j \defEq \ell_{\mathrm{snm}}( y_j, f( x_j ); \bar{\YCal}_j )$ with top-$\kLabel$ labels $\forall j$
            \STATE take gradient step with 
            $\mathrm{avg}_{\mathrm{top}( \kExample )}( \{ \ell_j \}_{j \in [ \MinibatchSize ]} )$
        \ENDFOR
    \end{algorithmic}
\end{algorithm}

\subsection{Expected loss under \SSNM{}}

Recall that in SGD, one works with the stochastic 
average minibatch loss
$\avgLoss( f; \Minibatch )$ per~\eqref{eqn:minibatch-loss}.
A key property of this loss is that its expectation is precisely the quantity we wish to optimise, namely, the empirical risk $\avgLoss( f; S )$.

In \SSNM{}, our minibatch loss~\eqref{eqn:minibatch-top-k}
has two sources of stochasticity,
as we draw both a minibatch \emph{and} a label sample.
Further, the minibatch loss only keeps the top-ranked loss values, which makes its expected behaviour more subtle.
As with SNM and average top-$\kExample$ SGD,
this expected loss
involves the \emph{order-weighted average}~\citep{Usunier:2009} of a vector,
$\mathrm{owa}( z; \theta ) \defEq \sum_{i \in [N]} \theta_i \cdot z_{[i]}$,
where $z_{[i]}$ denotes the $i$th largest element of the vector $z$.
We have the following.

\begin{lemma}
\label{lemm:expected-loss}
Pick any $\phi \colon \Real \to \Real_+$ 
and $\kExample$,
with induced SNM loss $\ellSNM$ per~\eqref{eqn:snm-bowl}.
Then, the expected \SSNM{} loss (cf.~ \eqref{eqn:minibatch-top-k}) over the draw of minibatch $\Minibatch$ and label set $\bar{\YCal}$ is
\begin{align*}
    &\E{}{ \widehat{L}_{\mathrm{top}( \kExample )}( f; \Minibatch ) } = \mathrm{owa}\left( \left\{ \bar{\ell}( y_i, f( x_i ) ) \colon i \in [ N ] \right\}; \theta \right) \\
    &{\bar{\ell}( y, f( x ) ) \defEq \phi( f_y ) +  \mathrm{owa}\left( \left\{ \phi( -f_{y'}( x ) ) \colon {y' \neq y} \right\}; \varphi \right)},
\end{align*}
where $\theta \in \Real_+^N, \varphi \in \Real_+^{\NumLabels - 1}$ are fixed weight vectors.
\end{lemma}

Concretely,
suppose $\kExample = 1$,
i.e.,
we compute the maximal loss over the minibatch.
By Lemma~\ref{lemm:expected-loss}, the expected loss 
is \emph{not} the maximum loss i \eqref{eqn:max-loss},
but rather an order-weighted average of the per-example losses.
We thus place a low, but non-zero, weight on smaller per-example losses.

We thus see that \SSNM{} involves a \emph{doubly order-weighted} average:
the expected loss emphasises the individual examples with high loss,
where the per-example loss in turn emphasises the individual labels with high loss. \citet{Reddi:2019} establish that the latter guarantees a tight bound on the retrieval performance.
In~\S\ref{sec:analysis}, we show that the former ensures good performance on subpopulations.

\subsection{Consistency of average top-$\kExample$ for retrieval}

A minimal requirement to impose
on the expected loss in Lemma~\ref{lemm:expected-loss}
is that it is \emph{consistent}~\citep{Zhang:2004,Bartlett:2006} for retrieval,
i.e.,
that its minimisation also ensures that the retrieval risk is minimised.
Here, we establish this when $\ell$ is the softmax cross-entropy, 
provided that the label distribution $\Pr_{\Y \vert \X}$ satisfies certain regularity conditions and $\kExample$ is chosen suitably as the number of sample $N$ increases. 

%
\begin{proposition}
\label{prop:consistency}
Pick the softmax cross-entropy loss $\ell$. 
Let $\alpha \defEq \lim_{N \to \infty} \frac{\kExample( N )}{N}$.
Then, 
for any $\alpha > 0$,
there exists 
an integer $r( \alpha )$
depending on $\Pr_{\Y \vert \X}$,
such that
the average top-$\kExample( N )$ loss with $\ell$
is consistent for top-$r(\alpha)$ retrieval.
\end{proposition}


%% file: analysis.tex
We now establish that minimising the average top-$\kExample$ loss controls 
performance
over heterogeneous \emph{subpopulations} of the data.
Thus, in retrieval settings, 
we ensure that good retrieval over 
different data subpopulations.

Formally, following~\S\ref{sec:challenges}, suppose the underlying data distribution comprises $P$ distinct subpopulations $\{\DCal_p\}_{p\in [P]}$;
i.e., $\DCal = \sum_{p \in [P]}\nu^{\ast}_{p} \cdot \DCal_p$ for some unknown mixture weights $\nu^{\ast} \in \Delta_P$.
Our aim is to control the following \emph{maximal retrieval loss} 
across different subpopulations:
\begin{align} 
\label{eq:max_retrieval_loss}
\max_{p \in [ P ]} {L}_{\mathrm{ret}( r )}( f; {\DCal}_p ),
\end{align}
where ${L}_{\mathrm{ret}( r )}$ is the retrieval loss for integer $r$ (cf.~\eqref{eqn:top-k-risk}).
We make two related comments here.
First, any $\DCal$ will admit an infinitude of  mixture representations involving different $\DCal_p$'s.
Second, performance on a given $\DCal_p$ ought to depend on how well-represented it is in $\DCal$,
i.e., the magnitude of $\nu^*_p$.
This intuition will be manifest in our subsequent bounds.

For simplicity, we focus on the full batch setting, i.e., $\MinibatchSize = N$.
We also consider the use of a generic multiclass loss $\ell$,
which could be the expected SNM loss from Lemma~\ref{lemm:expected-loss}.
Our analysis easily generalises to the minibatch setting,
using Lemma~\ref{lemm:expected-loss}
and the techniques of~\citet[Theorem 5]{Reddi:2019},~\citet[Theorem 2]{Kawaguchi:2019}.

%
\subsection{Warm-up: known subpopulation membership}
\label{sec:known-membership}

As a warm-up, 
we illustrate how we can control~\eqref{eq:max_retrieval_loss} if 
we know in advance which subpopulation each training example in $S$ belongs to.
For $p \in [P]$, let 
$\SCal_p \defEq \{i \in [N] : \XIYI \sim \DCal_p\}$ with $N_p:= |S_p|$ be the indices of the samples drawn from $\DCal_p$, the distribution of the $p$th subpopulation. 
Given a surrogate loss $\ell$, e.g., the SNM loss of~\eqref{eqn:snm-bowl},
the empirical loss for each subpopulation is
\begin{align}
\label{eq:empirical-sub}
{\widehat{L}_{\rm avg}(f; {\SCal}_p)} \defEq \frac{1}{N_p}\sum_{i \in \SCal_p}\ell( y_i, f( x_i ) )~~~\forall~p \in [P].
\end{align}
To achieve good performance on each subpopulation,
we may consider the \emph{agnostic} empirical loss from~\citet{Mohri:2019},
which considers the worst-case performance over \emph{all} mixtures of subpopulations:
for a mixture set $\Lambda \subseteq \Delta_P$,
\begin{align}
    \label{eq:empirical-agnostic}
    \widehat{L}_{\rm max}(f; S, \Lambda)
    \defEq \max_{\nu \in \Lambda} \sum_{p \in [P]} \nu_p \cdot {\widehat{L}_{\rm avg}(f; {\SCal}_p)}.
\end{align}
{Observe that if we pick 
$\Lambda = \Delta_P$,
then the maximum in \eqref{eq:empirical-agnostic} is exactly~\eqref{eq:max_retrieval_loss}.}
Equipped with this,
one may then appeal to generalisation bounds for the agnostic empirical loss~\citep[Theorem 2]{Mohri:2019}
to conclude that with high probability over the draw of the training sample,
\begin{align*}
    \max_{p \in [ P ]} {L}_{\mathrm{ret}( r )}( f; {\DCal}_p ) &\leq \widehat{L}_{\mathrm{max}}( f; S, \Lambda ) + \mathfrak{C}_{\FCal, \Lambda_{\mathrm{top}}, N}, 
\end{align*}
where $\mathfrak{C}_{\FCal, \Lambda_{\mathrm{top}}, N}$ depends on
a \emph{weighted} Rademacher complexity of the function class $\FCal$, 
and covering number of the mixture set $\Lambda_{\mathrm{top}}$;
see~\citet{Mohri:2019} for details.


{We now move
to the main
focus of this paper,} the more common and and challenging setting where the subpopulation membership for the training samples is \emph{unknown}. 
In this case, we cannot even compute the empirical loss over each subpopulation~\eqref{eq:empirical-sub}, let alone minimise the agnostic empirical loss~\eqref{eq:empirical-agnostic} to ensure good performance over each subpopulation. 
Fortunately, we can establish that
the proposed average top-$\kExample$ minimisation controls the loss in \eqref{eq:max_retrieval_loss}.
Our analysis proceeds as follows:
\begin{enumerate}[label=(\roman*),itemsep=0pt,topsep=0pt]
    \item first, we relate the 
    top-$\kExample$ risk that \SSNM{} minimises
    to the \emph{conditional value at risk} ({CVaR})
    
    \item next, we further relate 
    the 
    population version of the
    CVaR to the
    maximal subpopulation loss in~\eqref{eq:max_retrieval_loss}
    
    \item finally, we derive a generalisation bound for the CVaR, which shows that the maximal retrieval loss can be controlled by minimising the empirical top-$\kExample$ risk. 
\end{enumerate}

\subsection{From top-$\kExample$ to maximal retrieval loss via CVaR}

We begin by introducing
the \emph{conditional value at risk} (\emph{CVaR}) at level $\alpha \in (0, 1]$ of a random variable $\mathsf{U}$, 
defined as~\citep[Theorem 1]{Rockafellar:2000}
\begin{align}
\label{eq:cvar-def}
\cvar_{\alpha}( \mathsf{U} ) = \inf_{t \geq 0} \E{}{ \frac{1}{\alpha}[\mathsf{U}-t]_{+} + t },
\end{align}
where $[x]_{+} \defEq \max\{x, 0\}$.
Given iid samples $u_i \sim \mathsf{U}$,
we may construct an empirical estimate
\begin{equation}
\label{eqn:cvar-empirical}
\mathrm{CVaR}_{\alpha}( \hat{\mathsf{U}} ) \defEq 
\inf_{t \geq 0}\bigg\{\frac{1}{N\alpha}\sum_{i = 1}^{N}[u_i - t]_{+} + t\bigg\}.
\end{equation}
for $\alpha = \frac{k'}{N}$.
Here, $\mathrm{CVaR}_{\alpha}( \hat{\mathsf{U}} )$ is a \emph{biased} estimate of $\cvar_{\alpha}( \mathsf{U} )$,
since the expectation of the infimum is not the same as the infimum of the expectation~\citep{Brown:2007}.

Interestingly,~\eqref{eqn:cvar-empirical} is equivalent to the
average of the top-$\kExample$ largest elements of 
$\{ u_i \}_{i = 1}^N$
~\citep[Proposition 8]{Rockafellar:2002}, \citep[Lemma 2]{Fan:2017}.
This gives a means of re-expressing the average top-$\kExample$ loss in terms of the CVaR.
Specifically,
consider the random variable of loss values 
$\mathsf{L}_f \defEq \ell( \Y, f( \X ) )$ for $( \X, \Y ) \sim \DCal$
and given scorer $f$.
Given $N$ examples $S = \{ ( x_i, y_i ) \}_{i = 1}^N \sim \DCal^N$, we obtain $N$ measurements $\{ \ell( y_i, f( x_i ) ) \}_{i = 1}^N$ of the random variable $\mathsf{L}_f$.
Thus, for $\alpha = \frac{\kExample}{N}$, the empirical average top-$\kExample$ risk 
\begin{align}
\label{eq:emp-cvar-loss}
\widehat{L}_{\mathrm{top}( \kExample )}( f; S ) \defEq {\mathrm{CVaR}_{\alpha}( \widehat{\mathsf{L}}_f )},
\end{align}
gives us an empirical estimate of {$\mathrm{CVaR}_{\alpha}( {\mathsf{L}}_f )$}.

This connection between the empirical average top-$\kExample$ risk and CVar has twofold advantage.
First, the representation in~\eqref{eq:cvar-def} enables us to derive generalisation bounds for the retrieval problem. Second, appealing to a distinct representation for the CVaR 
evinces how it controls subpopulation performance~\citep[Example 4.2]{BenTal:2007}:
\begin{align*}
    \numberthis
    \label{eqn:cvar-subpopulation}
    \mathrm{CVaR}_{\alpha}( \mathsf{U} ) &= \sup_{Q \in \mathscr{B} ( P, \alpha )} \E{\mathsf{U}' \sim Q}{ \mathsf{U}' }, 
\end{align*}
where the uncertainty set $\mathscr{B} ( P, \alpha ) = \{ Q \in \Delta_{\UCal} \mid (\exists R \in \Delta_{\UCal}) \, P = \alpha \cdot Q + (1 - \alpha) \cdot R \}$ with $\Delta_{\UCal}$ denoting the set of all distributions on the sample space $\UCal$.
The CVaR is thus the maximal expectation under any \emph{minority subpopulation} of the original distribution
with mass at least $\alpha$.
A similar connection was made in~\citep[Eq. (2)]{Duchi:2019}.

{{The representation in~\eqref{eqn:cvar-subpopulation} can be seen as a generalisation of the agnostic loss in~\eqref{eq:empirical-agnostic},
wherein there are \emph{subpopulation dependent} mixing weights,
and the empirical subpopulation losses are replaced with average top-$\kExample$ counterparts.}

\begin{lemma}
\label{lemm:top-k-agnostic}
{Pick a scorer $f$
and $\kExample \in \mathbb{N}_+$.
For any sample $S$ with subpopulation indices $\{ \SCal_p \}_{p \in [ P ]}$ of sizes $N_p \defEq | \SCal_p |$,}
\begin{align}
    \label{eq:agnostic-topk}
    &
    \mathrm{CVaR}_{\frac{\kExample}{N}}( \widehat{\mathsf{L}}_f ) =
    \max_{\nu \in \TCal_{k',p}} \sum_{p \in [P]} \nu_p \cdot {\avgTop{k' \nu_p}{\SCal_p}},
    \\
    &
    \TCal_{k',p} \defEq \Big\{\nu \in \big\{0, {1}/{k'},\ldots, 1\big\}^P : k' \nu_p \leq N_p, \sum_{p \in [P]}\nu_p = 1\Big\}
    .
    \nonumber
\end{align}
\end{lemma}
}
We may thus relate the top-$\kExample$ risk
to a maximal risk over heterogeneous subpopulations.
Our next step is to convert this to a generalisation bound
for empirical top-$\kExample$ minimisation.

\subsection{Generalisation bounds for CVaR and top-$\kExample$}

From the above,
the key to deriving a generalisation bound is to study the behaviour of the CVaR.
{Specifically, 
how well does the \emph{empirical} quantity $\mathrm{CVaR}_{\alpha}( \widehat{\mathsf{L}}_f )$
approximate the \emph{population} CVaR on ${\mathsf{L}}_f$, i.e., $\mathrm{CVaR}_{\alpha}( {\mathsf{L}}_f )$?}
The first concern is that $\mathrm{CVaR}_{\alpha}( \widehat{\mathsf{L}}_f )$ is \emph{not} an unbiased estimate of $\mathrm{CVaR}_{\alpha}( {\mathsf{L}}_f )$.
Fortunately, this bias asymptotically vanishes.

\begin{lemma}
\label{lemm:cvar-bias}
Pick any $\alpha \in (0, 1)$.
Fix a function class $\FCal \subset \Real^{\XCal}$
and
bounded
loss $\ell \colon [ \NumLabels ] \times \Real^{\NumLabels} \to \Real_+$.
Then, 
for every $f \in \FCal$,
$$ \mathrm{CVaR}_{\alpha}( \mathsf{L}_f ) - \mathbb{E}[ \mathrm{CVaR}_{\alpha}( \widehat{\mathsf{L}}_f ) ] = \mathscr{O}\left( N^{-1/2} \right). $$
\end{lemma}

We may thus seek a generalisation bound for $\mathrm{CVaR}_{\alpha}( \widehat{\mathsf{L}}_f )$ in terms of its population counterpart,
with an additional bias term that asymptotically vanishes.
While concentration bounds for the CVaR are well-studied~\citep{Brown:2007,WangGao:2010},
these do not suffice for our purposes since 
we seek a \emph{uniform} bound over all $f \in \FCal$.

The following result presents two such bounds. The first bound relies on a result of~\citet{Cherukuri:2019},
and depends on the covering number of $\FCal$. The second bound is an application of Rademacher bounds to the formulation in~\eqref{eq:cvar-def}.
\begin{proposition}
\label{prop:cvar-gen-bound}
Pick any $\alpha, \delta \in (0, 1)$.
Fix a function class $\FCal \subset \Real^{\XCal}$
with diameter $\mathrm{diam}( \FCal )$,
and bounded,
1-Lipschitz loss $\ell \colon [ \NumLabels ] \times \Real^{\NumLabels} \to \Real_+$
with $B \defEq \| \ell \|_{\infty}$.
Then, with probability at least $1 - \delta$ over $S \sim \DCal^N$,
for every $f \in \FCal$,
\begin{align*}
    &\mathrm{CVaR}_{\alpha}( \mathsf{L}_f ) \leq \mathrm{CVaR}_{\alpha}( \widehat{\mathsf{L}}_f ) + (M_1 \land M_2) + \mathscr{O}\left( N^{-1/2} \right), 
\end{align*}
where 
\begin{align*}
    M_1 &\defEq \OCal\left(\min\bigg\{\frac{\mathrm{diam}(\FCal)}{\alpha\delta}, B\sqrt{\frac{1}{\alpha N}\log\bigg(\frac{\mathrm{diam}(\FCal)}{\alpha \delta}\bigg)}\bigg\}\right) \\
    M_2 &\defEq  \frac{1}{\alpha} \cdot \left( \mathrm{Rad}_N( \FCal ) + \frac{B}{\sqrt{N}} \right) + \sqrt{\frac{\log ({1}/{\delta})}{2N}},
\end{align*}
for
empirical Rademacher complexity
$\mathrm{Rad}_{N}( \FCal )$.
\end{proposition}

An immediate consequence of
Proposition~\ref{prop:cvar-gen-bound}
is that minimising the 
\emph{empirical top-$\kExample$} risk
for $\kExample = \alpha N$
controls
the \emph{maximum subpopulation risk}:
using~\eqref{eq:cvar-def}, \eqref{eq:emp-cvar-loss}, and
~\eqref{eqn:cvar-subpopulation},
\begin{align*}
    &\sup_{\DCal' \in \mathscr{B} ( \DCal, \alpha )} \E{( \X, \Y ) \sim \DCal'}{ \ell( \Y, f( \X ) ) } 
    \leq \widehat{L}_{\mathrm{top}( \kExample )}( f; S ) +( M_1 \land M_2 ) + \mathscr{O}\left( N^{-1/2} \right),
\end{align*}
where the uncertainty set $\mathscr{B} ( \DCal, \alpha )$ comprises all subpopulations of $\DCal$ occupying at least $\alpha$ mass. The LHS is exactly the maximal retrieval loss in~\eqref{eq:max_retrieval_loss}, assuming $\DCal = \sum_{p \in [P]} \nu^*_p \cdot \DCal_p$ with $\nu^*_p \geq \alpha$, for all $p \in [P]$.
This is intuitive:
as the subpopulation becomes rarer ($\alpha \to 0$), it is harder to guarantee good performance using a finite number of samples ($M_1, M_2 \to \infty$).

\subsection{Relation to existing work}

The two key steps in our analysis are to relate the top-$k$ loss to the CVaR,
and to then relate the CVaR to a worst-case subpopulation loss.
The implication of combining these observations for retrieval has not previously been noted.

The issue of controlling the risk across subpopulations has been studied in prior work.
\citet{Shalev-Shwartz:2016} showed that minimising the maximal loss over all examples
 can guarantee good performance on data sub-populations.
However, this analysis was for
separable binary problems.
By contrast, our analysis is for non-separable multiclass problems, and uses the more robust top-$\kExample$ loss.
\citet{Fan:2017} proposed to minimise the average of the top-$\kExample$ largest per-example losses.
However,
their analysis only established
consistency in binary classification settings,
and did not establish guarantees on different subpopulations.

In the fairness literature,
a line of work has explored the use of the maximum or average top-$\kExample$ per-subgroup losses~\citep{Alabi:2018,Mohri:2019,Williamson:2019}.
These assume the subgroup memberships are known,
akin to~\S\ref{sec:known-membership},
which is not the case in our setup.

When subgroup membership is unknown,~\citet{Duchi:2018,Hashimoto:2018,Oren:2019,Duchi:2019} propose robust optimisation procedures.
These can be seen as directly working with 
variational representations akin to~\eqref{eq:cvar-def},
which requires 
fundamentally altering the training procedure
(e.g., introducing an auxiliary parameter to be iteratively optimised over);
further, such techniques are not attuned to the retrieval setting, where $K$ is large.
By contrast, 
the 
minibatch top-$\kExample$ minimisation in
\SSNM{} requires only a minor modification to the {state-of-the-art} SNM retrieval method~\citep{Reddi:2019}.

%% file: experiments.tex
We now present experiments confirming our central claim that \SSNM{} can 
ensure good performance on heterogeneous data subpopulations,
while being efficient to optimise in settings with a large number of labels.
We will illustrate this on both a controlled setting where the subpopulations are artificially created,
and a setting with natural subpopulations.

At the outset, we emphasise 
that 
to our knowledge, 
\emph{no} existing retrieval method addresses the hetereogenity problem;
state-of-the-art retrieval methods~\citep{Jain:2016,Prabhu:2018,Reddi:2019,Guo:2019}
focus on aggregate performance, which we shall demonstrate can lead to a bias towards a dominant subpopulation.
Further, while we do \emph{not} claim to improve upon the state-of-the-art in terms of standard retrieval performance
{--- involving metrics agnostic to heterogeneous subpopulations --- }
we show that \SSNM{} remains competitive on such metrics as well.



%
\textbf{Results on small-scale data}.
We begin with an experiment 
on two benchmark datasets, \mnist~and \cifarsmall,
each of which has $10$ classes.
On each dataset,
we artificially create pronounced subpopulations as follows:
\begin{enumerate}[label=(\alph*),itemsep=-2pt,topsep=-2pt]
    \item we split the classes into two groups, Head \& Tail
    \item on the training set, we downsample the Tail classes such that the ratio of Head$\colon$Tail classes is $100 \colon 1$.
\end{enumerate}
The data associated with the resulting Head and Tail classes form two imbalanced subpopulations of the data distribution. 
We do \emph{not} modify the test data in any way,
and so there is no such population imbalance in the test set.

We consider standard neural architectures for these datasets:
for \mnist, we train a \lenet~\citep{LeCun:1998} with minibatch size $\MinibatchSize = 64$ for $100K$ iterations,
and
for \cifarsmall, we train a \resnet-50~\citep{He:2016} with batch size $\MinibatchSize = 64$ for $600K$ iterations.
On each dataset,
we train \SSNM{} for $\kExample \in \{ 1, 16, 32, 64 \}$,
with softmax cross-entropy as our base loss.
Since the number of labels in both datasets is only $\NumLabels = 10$, we do not perform sampling over labels, i.e., $\kLabel = K$.
Recall from~\S\ref{sec:special} that this corresponds to performing average top-$\kExample$ SGD and
that when $\kExample = 64$, the method is identical to standard minibatch SGD.

\begin{table}[!t]
    \centering
    \begin{tabular}{@{}lllll@{}}
        \toprule
        \textbf{Dataset} & \textbf{Method} & \textbf{Head} & \textbf{Tail} & \textbf{Full} \\
        \toprule
        \multirow{4}{*}{MNIST} & Top-$1$   & 0.9521 & 0.4412 & 0.6915 \\
        & Top-$16$  & 0.9956 & \best{0.8826} & \best{0.9380} \\
        & Top-$32$  & 0.9959 & 0.8636 & 0.9288 \\
        & All $64$ (SGD) & \best{0.9964} & 0.8736 & 0.9341 \\
        \midrule
        \multirow{4}{*}{CIFAR-10} & Top-$1$ & 0.5273 & 0.3657 & 0.4465 \\
        & Top-$16$  & 0.8656 & 0.4743 & 0.6700 \\
        & Top-$32$  & \best{0.8734} & \best{0.4875} & \best{0.6805} \\
        & All $64$ (SGD)  & 0.8594 & 0.4694 & 0.6645 \\
        \bottomrule
    \end{tabular}
    \caption{Test set accuracy on \mnist~and \cifarsmall, where the training set is downsampled so that $5$ classes appear 1\% of the time. We assess average top-$\kExample$ minibatch SGD for varying $\kExample$, on a minibatch size of $\MinibatchSize = 64$.
    The best performance on the tail subpopulation defined by the downsampled classes is achieved by choosing $\kExample < 64$.
    }
    \label{tbl:results-mnist}
\end{table}

Table~\ref{tbl:results-mnist} shows the performance of all methods on both the downsampled and unaltered classes. We see that on both datasets, 
using the average top-$\kExample$ loss for $\kExample < 64$ results in the best performance on {\em tail}, i.e., the rare subpopulation corresponding to the downsampled classes. Further, on the \cifarsmall~dataset, we even see slight gains on modelling {\em head}, i.e., the subpopulation associated with the unaltered classes.
This indicates that even within the head, we are able to better model more difficult examples.



%
\textbf{Results on large-scale data}.
We next consider large-scale classification on \amazonsmall,
a standard benchmark for extreme \emph{multilabel} classification. Following~\citet{Reddi:2019}, we convert this to a multiclass dataset as follows: given an instance $x$ and its $m$ positive labels $\{y_1,\ldots, y_m\} \subseteq [\NumLabels]$ in the original dataset, we add $m$ multiclass examples $\{(x, y_1),\ldots, (x, y_m)\}$ in the new dataset. 

Unlike the previous experiment,
on this dataset we consider the \emph{naturally} occurring subpopulations induced by the inherent label imbalance.
We thus do \emph{not} modify the training set in any way;
rather, at test time, we simply evaluate performance \emph{per-subpopulation}.
These subpopulations are defined based on the label frequency as follows:
\begin{enumerate}[label=(\alph*),itemsep=0pt,topsep=0pt]
    \item for each label $y \in [\NumLabels]$, we compute the fraction $\pi_y$ of times it appears in the training set
    
    \item we let $q, q'$ be the 66\% and 33\% quantile of the $\pi_y$'s

    \item we say
    a test sample $(x, y)$ belongs to Head if $\pi_y > q$,
Torso if $q \geq \pi_y > q'$,
and Tail otherwise.
\end{enumerate}
At test time, we then separately evaluate performance on each of these three subgroups.
Observe that by this construction, the subpopulations have \emph{overlapping support}:
indeed, an instance $x$ may be associated with both rare and frequent labels, in which case it will belong to multiple subgroups.

To measure performance on a subpopulation, 
we compute recall@$r$ for varying values of $r \in \{ 5, 10, 25, 50 \}$, 
which is simply the negation of the retrieval risk~(cf.~\eqref{eqn:top-k-risk}).
This measures the number of $(x, y)$ pairs where $y$ is amongst the top-$r$ scored labels from the model.
Since each instance can have many labels and frequent labels tend to get higher scores compared to rare labels, we expect higher recall@$r$ for Head labels, especially for low values of $r$.  

\begin{table}[!t]
    \centering
    \begin{tabular}{@{}lllllll@{}}
        \toprule
         \textbf{Metric} & \textbf{Method} & \textbf{Head} & \textbf{Torso} & \textbf{Tail} & \textbf{Full} \\
        \toprule
         \multirow{3}{*}{{\rm recall}@5} & Top-$256$  & 0.8143 & 0.5302 & 0.4591 & 0.5822 \\
         & Top-$512$  & \best{0.8271} & 0.5347 & 0.4682 & 0.5911 \\
         & All $2048$ (SNM)  & 0.8241 & \best{0.5475} & \best{0.4944} & \best{0.6115} \\
        \cmidrule{2-6}
         \multirow{3}{*}{{\rm recall}@10} & Top-$256$  & 0.9081 & 0.7005 & 0.6205 & 0.7313 \\
         & Top-$512$  & \best{0.917} & \best{0.7111} & 0.6388 & 0.7358 \\
          & All $2048$ (SNM)  & 0.9016 & 0.7041 & \best{0.6429} & \best{0.7419} \\
        \cmidrule{2-6}
         \multirow{3}{*}{{\rm recall}@25} & Top-$256$  & 0.9716 & 0.8556 & 0.7775 & 0.8632 \\
         & Top-$512$  & \best{0.9762} & \best{0.8718} & \best{0.7885} & \best{0.8661} \\
         & All $2048$ (SNM) & 0.9507 & 0.8313 & 0.7681 & 0.8439\\
        \cmidrule{2-6}
         \multirow{3}{*}{{\rm recall}@50} & Top-$256$  & 0.9866 & 0.9319 & 0.8613 & 0.9215 \\
        & Top-$512$  & \best{0.9896} & \best{0.9353} & \best{0.8629} & \best{0.9254} \\
        &  All $2048$ (SNM) & 0.9684 & 0.8974 & 0.8428 & 0.8913 \\
        \bottomrule
    \end{tabular}
    \caption{Test set recall on \amazonsmall, where the evaluation is split into three subpopulations (Head, Torso, Tail) based on label popularity. We assess \SSNM{} for varying values of $\kExample$, on a minibatch size of $\MinibatchSize = 2048$. Note that SNM corresponds to choosing $\kExample = 2048$.    
    During each optimization step, we select the $k = 64$ hardest negative labels from a set of $\bar{\NumLabels} = 4096$ randomly sampled negative labels. The best performance at high levels of recall is achieved by choosing $\kExample < 2048$.
    }
    \label{tbl:results-multilabel}
\end{table}

For minibatch size $\MinibatchSize = 2048$, 
we compare the performance of \SSNM{} for $\kExample \in \{ 256, 512, 2048 \}$.
Note that $\kExample = 2048$ refers to the baseline of SNM, which averages the loss over all examples whithin a batch.
Following~\citet{Reddi:2019}, we train a fully connected neural network with a single hidden layer of width $512$ employing linear activations. 
We use the cosine contrastive loss~\citep{Hadsell:2006} as our base loss $\ell$. 
During each step of training, for each instance, we select the $\kLabel = 64$ hardest negative labels from a set of $\bar{\NumLabels} = 4096$ negative labels sampled uniformly.

Table~\ref{tbl:results-multilabel} shows the retrieval performance across multiple recall thresholds on all subpopulations (Head, Torso, Tail), as well as the Full test set. 
From the results, we make the following key observations. 
First, on the Full test set, with increasing values of $r$ the benefit of the proposed method becomes more pronounced:
for $r=25$, we observe best performance by selecting the Top-512 examples per batch. 

Second, while we already observe increased recall of head labels in the top-5 results, with increasing $r$, recall further improves over the baseline across subpopulations of increasingly rare labels. 
For $r \geq 25$, performance improves across all subpopulations. 
This indicates that in addition to overall improved recall, \SSNM{} learns a consistent ranking across labels, i.e., head labels are consistently ranked within the top labels, followed by torso and then tail labels. 

Overall, \SSNM{} achieves good overall recall performance as well as across different data subpopulations, 
while being efficient to optimise with a large number of labels.

%% file: conclusion.tex
{We proposed {doubly-stochastic mining} (\SSNM{}),
wherein we minimise the loss of the hardest examples \emph{and} labels.
Theoretically and empirically, \SSNM{} controls the retrieval loss over heterogeneous data subpopulations. 
Mitigating the effect of noise on the top-$\kExample$ loss, e.g. using ``semi-hard'' examples~\citep{Schroff:2015}, would be of interest.}

%% file: proofs.tex
\begin{proof}[Proof of Lemma~\ref{lemm:expected-loss}]
Let $\mathcal{N}_{\rm mb} \subseteq [N]$ denote the indices of the $N_{\rm mb}$ samples selected in the minibatch $\Minibatch$. Note that 
\begin{align}
\label{eq:lemma1-step1}
\E{\{\bar{\YCal}_i\},~\mathcal{N}_{\rm mb}}{ \avgTopK( f; \Minibatch ) } &= \E{\{\bar{\YCal}_i\}}{ \E{\mathcal{N}_{\rm mb}}{\avgTopK( f; \Minibatch ) \Big\vert \{\bar{\YCal}_i\} }} \nonumber \\
& \overset{(i)}{=} \E{\{\bar{\YCal}_i\}}{ \mathrm{owa}\left( \left\{  \ellSNM( y_i, f( x_i ); \bar{\YCal}_i ) \colon i \in [ N ]  \right\}; \theta \right)} \nonumber \\
& = \mathrm{owa}\left( \left\{\E{\bar{\YCal}_i}{   \ellSNM( y_i, f( x_i ); \bar{\YCal}_i )} \colon i \in [ N ]  \right\}; \theta \right),
\end{align}
where $(i)$ follows from~\citet{Kawaguchi:2019}. The precise form of the weights vector $\theta \in \Real_+^N$ is per~\citet{Kawaguchi:2019}.
Compared to SGD, the expected loss is a \emph{weighted} average of the loss over the samples. Further, the weighting places emphasis on samples with highest loss.
Next, by following~\citet{Reddi:2019}, we have
\begin{align}
\label{eq:lemma1-step2}
\E{\bar{\YCal}_i}{ \ellSNM( y_i, f( x_i ); \bar{\YCal}_i ) } = \phi( f_{y_i} ) + \mathrm{owa}\left( \{ \phi( -f_{y'} ) \}_{y' \neq y_{i}}; \varphi \right),
\end{align} 
where $\varphi \in \Real_+^{\NumLabels - 1}$ is a weight vector.
Here, the weighting places emphasis on the labels with highest loss.
(The precise form of the weights is per~\citet{Reddi:2019}.)
Now, Lemma~\ref{lemm:expected-loss} follows by combining \eqref{eq:lemma1-step1} and \eqref{eq:lemma1-step2}.
\end{proof}

\begin{proof}[Proof of Proposition~\ref{prop:consistency}]
We follow the proof of~\citet{Fan:2017}, which was done for the binary classification setting.
By exploiting the connection between the average top-$\kExample$ risk and the CVaR~\eqref{eq:cvar-def},
we have that
\begin{align}
\label{eq:consistency-step1}
\avgTopKPrime(f; S) \xrightarrow[N \to \infty]{\frac{k'}{N} = \alpha} \pAvgTopKPrime( f; \DCal ) = \inf_{t \geq 0} \E{(\X, \Y) \sim \DCal}{ \frac{1}{\alpha}[ \ell( \Y, f( \X ) ) - t ]_{+} + t }
\end{align}
where $\alpha = \frac{k'}{N}$. Let $(f^*, t^*)$ be the minimizer of the RHS in \eqref{eq:consistency-step1}.

Further, letting $f^*_{\ell}$ be the minimiser of the risk with respect to loss $\ell$, i.e., $f^*_{\ell} = \argmin_{f \colon \XCal \to \Real}\E{(\X, \Y) \sim \DCal}{\ell( \Y, f( \X )}$. Therefore, we have
\begin{align*}
     t^* &\leq \underbrace{\mathbb{E}\Big[ \frac{1}{\alpha}[ \ell( \Y, f^*( \X ) ) - t^* ]_{+}\Big]}_{\geq 0} +~ t^* = \E{}{ \frac{1}{\alpha}[ \ell( \Y, f^*( \X ) ) - t^* ]_{+} + t^* } \\
     &= \min_{f, t} \E{(\X, \Y) \sim \DCal}{ \frac{1}{\alpha}[ \ell( \Y, f( \X ) ) - t ]_{+} + t } \\
     &\leq \min_{t} \E{(\X, \Y) \sim \DCal}{ \frac{1}{\alpha}[ \ell( \Y, f^*_{\ell}( \X ) ) - t ]_{+} + t } \\
     &\leq \E{(\X, \Y) \sim \DCal}{ \frac{1}{\alpha}[ \ell( \Y, f^*_{\ell}( \X ) ) ]_{+} } \\
     &= \E{(\X, \Y) \sim \DCal}{ \frac{1}{\alpha} \cdot \ell( \Y, f^*_{\ell}( \X ) ) } \\
     \numberthis
     \label{eqn:t-leq-alpha}
     &= \frac{1}{\alpha} \cdot L^*,
\end{align*}
where $L^* = \min_{f \colon \XCal \to \Real}\E{(\X, \Y) \sim \DCal}{\ell( \Y, f( \X ) }$ denotes the Bayes-risk (minimal loss) with respect to $\ell$.

Equipped with this,
let us consider the following problem
\begin{align}
\label{eq:consistency-opt-prob}
&{\rm minimize}\quad \mathbb{E}_{Y\vert X = x}\left[\ell(Y, f(x)\right] = \sum_{y}p_{y}(x)\cdot[-\log g_y(x) - t]_{+} = \sum_{y}p_y(x)\cdot [-\log \tau g_y(x)]_{+} \nonumber \\
& {\rm subject~to}\quad \sum_{y}g_y(x) = 1~~\text{and}~~g_y(x) \geq 0~\forall y,
\end{align}
where $g_y(x) = \frac{e^{f_y(x)}}{\sum_{j}e^{f_j(x)}}$ and $\tau = e^{t}$. In what follows, we suppress the dependence on $x$ to obtain the following problem.
\begin{align}
\label{eq:consistency-opt-prob-1}
&{\rm minimize}\quad \sum_{y}p_y\cdot [-\log \tau g_y]_{+} \nonumber \\
& {\rm subject~to}\quad \sum_{y}g_y = 1~~\text{and}~~g_y \geq 0~\forall y,
\end{align}
The problem can be rewritten as 
\begin{align}
&{\rm min}_{g_y \geq 0~\forall~y}~{\rm \max}_{\lambda \in \Real} \sum_{y}p_y\cdot [-\log \tau g_y]_{+} + \lambda\cdot(1 - \sum_{y}g_y) - \sum_{y}\mu_y\cdot g_y \nonumber \\
&\qquad \geq {\rm \max}_{\lambda \in \Real}~{\rm min}_{g_y \geq 0~\forall~y} \underbrace{\sum_{y}p_y\cdot [-\log \tau g_y]_{+} + \lambda\cdot(1 - \sum_{y}g_y) - \sum_{y}\mu_y\cdot g_y}_{W}.
\end{align}
Note that 
\begin{align}
\label{eq:consistency-subgrad}
\frac{\partial W}{\partial g_y} = \begin{cases}
- \frac{p_y}{g_y} - \lambda & \text{if $-\log \tau g_y > 0$ or $g_y < \frac{1}{\tau}$} \\
\alpha_yp_y - \lambda~\text{for}~\alpha_y \in [0, 1] & \text{if $-\log \tau g_y = 0$ or $g_y = \frac{1}{\tau}$} \\
-\lambda & \text{if $-\log \tau g_y \leq 0$ or $g_y > \frac{1}{\tau}$}
\end{cases}
\end{align}
Let $\YCal_1 = \{y~:~g_y = \frac{1}{\tau}\}$ and $\YCal_2 = \{y~:~g_y < \frac{1}{\tau}\}$. Assuming that $\NumLabels \cdot \frac{1}{\tau} > 1$, we claim that $\YCal_1 \cup \YCal_2 = \YCal$, i.e., at the optimal solution we don't have $g_y > \frac{1}{\tau}$. This easily follows from the observation that assigning more that $\frac{1}{\tau}$ mass to $g_y$ does not further reduce the objective, however, it can increase the contribution of $g_{y'}$, for $y' \neq y$, to the objective. From KKT condition, an optimal solution satisfies
\begin{align}
\frac{\partial W}{\partial g_y} = 0~~\forall y
\end{align}
or
\begin{align}
g^{*}_y = -\frac{p_y}{\lambda^{*}}~~\forall~y \in \YCal_2.
\end{align}
Now, since $\{g_y\}$ forms a valid distribution, we have
\begin{align}
1 = \sum_{y}g^{*}_y = \sum_{y \in \YCal_1}g^{*}_y + \sum_{y \in \YCal_2}g^{*}_y = |\YCal_1|\cdot\frac{1}{\tau} - \sum_{y \in \YCal_2}\frac{p_y}{\lambda^{\ast}} = |\YCal_1|\cdot\frac{1}{\tau} - \frac{1}{\lambda^{*}}\cdot(1 - \sum_{y \in \YCal_1}p_y)
\end{align}
or 
\begin{align}
-\frac{1}{\lambda^{*}} = \frac{\left(1 - {|\YCal_1|}/{\tau}\right)}{(1 - \sum_{y \in \YCal_1}p_y)}.
\end{align}
Thus, we have, 
\begin{align}
g^{*}_y = \begin{cases}
\frac{1}{\tau} & y \in \YCal_1, \\
\frac{\left(1 - {|\YCal_1|}/{\tau}\right)}{(1 - \sum_{y \in \YCal_1}p_y)}\cdot p_y & y \in \YCal_2.
\end{cases}
\end{align}
Recall that, for $y \in \YCal_2$, we have $g^{*}_y  <  \frac{1}{\tau}$. Thus, $y \in \YCal_2$ if
\begin{align}
p_y < \frac{1}{\tau}\cdot\frac{(1 - \sum_{y \in \YCal_1}p_y)}{\left(1 - {|\YCal_1|}/{\tau}\right)}.
\end{align}

Let $\tau^* = e^{t^*}$.
Consequently,
the minimiser of the average top-$\kExample$ risk
collapses the probabilities of the largest $| \YCal_1 |$ labels into $\frac{1}{\tau^*}$.
Further, these largest elements will have probability larger than the other labels.
Consequently,
the minimiser will be
will be top-$\tau^*$ consistent.

To get some intuition,
recall that $\tau^* = e^{t^*}$ for the optimal threshold $t^*$.
Thus, in light of~\eqref{eqn:t-leq-alpha},
$\tau^* \leq e^{L^*/\alpha}$.
When $L^* \sim 0$ (i.e., the problem is noise-free), 
and $\alpha \sim 1$ (i.e., we average most of the labels),
this indicates that the loss will be top-$1$ consistent, as is intuitive.
\end{proof}

\begin{proof}[Proof of Lemma~\ref{lemm:top-k-agnostic}]
Consider the following set of weightings on the training samples:
\begin{equation}
\label{eq:k-sparse}
\Lambda_{\mathrm{top}( \kExample ), N} \defEq \{\tau \in \{0, 1/k'\}^N : \|\tau\|_0 = k' \} \subset \Delta_N,
\end{equation}
i.e.,
all $k'$-sparse vectors with equal weight on the nonzero components.
The empirical average {\rm top}-$k'$ loss as defined in \eqref{eqn:minibatch-top-k} can then be rewritten as 
\begin{align}
\label{eq:atopk-a}
\avgTopKPrime( f; S ) = \max_{\tau \in \Lambda_{\mathrm{top}( \kExample ), N}}\sum_{i \in [N]} \tau_i \cdot \ell(y_i, f( x_i )).
\end{align}
Given $\tau \in \Lambda_{\mathrm{top}( \kExample ), N}$, 
for each subpopulation,
we can further consider only those samples which have nonzero associated weight:
$ \SCal_{\tau,p} = \{i \in \SCal_p : \tau_i > 0 \} \subseteq \SCal_p, p \in [P].$
Now, we can rewrite \eqref{eq:atopk-a} as follows:
\begin{align*}
&\avgTopKPrime( f; S ) = \max_{\tau \in \Lambda_{\mathrm{top}( \kExample ), N}} \sum_{p \in [P]} \frac{1}{k'} \sum_{i \in \SCal_{\tau,p}} \ell(y_i, f( x_i )) \\
&\qquad= \max_{\tau \in \Lambda_{\mathrm{top}( \kExample ), N}} \sum_{p \in [P]} \frac{|\SCal_{\tau,p}|}{k'} \cdot \frac{1}{|\SCal_{\tau,p}|} \sum_{i \in \SCal_{\tau,p}} \ell(y_i, f( x_i )).
\end{align*}
Note that the maximisation over $\tau$ is equivalent to selecting $\lambda_p \defEq \frac{| \SCal_{\tau, p} |}{k'}$ fraction of the $k'$ hardest examples that define the average top-$\kExample$ loss from the $p$-th subpopulation. We may thus write
\begin{align*}
    &
    \avgTopKPrime( f; S ) = \max_{\lambda \in \TCal_{k',p}} \sum_{p \in [P]} \lambda_p \cdot {\avgTop{k' \lambda_p}{\SCal_p}}, \\
    &
    \TCal_{k',p} \defEq \Big\{\lambda \in \big\{0, {1}/{k'},\ldots, 1\big\}^P : k' \lambda_p \leq N_p, \sum_{p \in [P]}\lambda_p = 1\Big\}.
\end{align*}
\end{proof}

\begin{proof}[Proof of Lemma~\ref{lemm:cvar-bias}]
Fix any $f \in \FCal$.
From the definition of the empirical and population CVaR~\eqref{eqn:cvar-empirical},~\eqref{eq:cvar-def}, 
we have
\begin{align*}
    \E{}{ \mathrm{CVaR}( \mathsf{L}_f ) - \mathrm{CVaR}( \hat{\mathsf{L}}_f ) }
    &= \E{}{ \frac{1}{\alpha} \cdot [ \ell( \Y, f( \X ) ) - t^* ]_+ + t^* } - \E{}{ \frac{1}{N \alpha} \sum_{i = 1}^{N} [ \ell( y_i, f( x_i ) ) - \hat{t}^* ]_+ + \hat{t}^* },
\end{align*}
where $\hat{t}^*, t^*$ are the optimal values of $t$ in~\eqref{eqn:cvar-empirical},~\eqref{eq:cvar-def},.
Following~\citet{Rockafellar:2000},
these correspond to the empirical and population quantiles of the losses, which we denote by $\hat{q}_{\alpha}$ and $q_{\alpha}$ respectively.
Thus,
\begin{align*}
    \E{}{\mathrm{CVaR}( \mathsf{L}_f ) - \mathrm{CVaR}( \hat{\mathsf{L}}_f )}
    &= \E{}{ \frac{1}{\alpha} \cdot [ \ell( \Y, f( \X ) ) - q_\alpha ]_+ } - \E{}{ \frac{1}{N \alpha} \sum_{i = 1}^{N} [ \ell( y_i, f( x_i ) ) - \hat{q}_{\alpha} ]_+ } + \E{}{ (q_\alpha - \hat{q}_{\alpha}) } \\
    &= \frac{1}{N \alpha} \sum_{i = 1}^{N} \E{}{ [ \ell( y_i, f( x_i ) ) - q_\alpha ]_+ } - \E{}{ \frac{1}{N \alpha} \sum_{i = 1}^{N} [ \ell( y_i, f( x_i ) ) - \hat{q}_{\alpha} ]_+ } +  \E{}{ (q_\alpha - \hat{q}_{\alpha}) } \\
    &= \frac{1}{N \alpha} \sum_{i = 1}^{N} \E{}{ [ \ell( y_i, f( x_i ) ) - q_\alpha ]_+ - [ \ell( y_i, f( x_i ) ) - \hat{q}_{\alpha} ]_+ } + \E{}{ (q_\alpha - \hat{q}_{\alpha}) }.
\end{align*}
A simple case-analysis reveals that $\left| [ \ell( y_i, f( x_i ) ) - q_\alpha ]_+ - [ \ell( y_i, f( x_i ) ) - \hat{q}_{\alpha} ]_+ \right| \leq | q_\alpha - \hat{q}_{\alpha} |$.
Thus, the RHS can be bounded by $\left( 1 + \frac{1}{\alpha} \right) \cdot \E{}{ | q_\alpha - \hat{q}_\alpha | }$.
Now observe that the empirical quantile $\hat{q}_\alpha$ converges in distribution to a Gaussian with mean $q_{\alpha}$ and variance $\OCal\left( \frac{1}{{N}} \right)$~\citep[Theorem 8.5.1]{Arnold:1992}.
Since the mean of the absolute value of a zero-mean Gaussian with variance $\sigma^2$ is $\OCal( \sigma )$, the result follows.
\end{proof}

\begin{proof}[Proof of Proposition~\ref{prop:cvar-gen-bound}]
We provide both a covering number bound
based on~\citet{Cherukuri:2019},
and a Rademacher bound based on~\citet{Boyd:2012}.

\textbf{Covering number bound}.
Note that for any fixed $f \in \FCal$,
\begin{align}
\label{eq:step0}
\mathrm{CVaR}( \mathsf{L}_f ) - \mathrm{CVaR}( \hat{\mathsf{L}}_f ) \leq \sup_{h \in \FCal} \mathrm{CVaR}( \mathsf{L}_h ) - \mathrm{CVaR}( \hat{\mathsf{L}}_h ) \Big).
\end{align} 
Let the loss function $\ell : [ \NumLabels ] \times \Real^{\NumLabels} \to \Real$ satisfy the Lipschitz condition:
\begin{align}
\|\ell(h; z) - \ell(h'; z)\|_2 \leq \gamma \|h - h' \|_2.
\end{align}
Under this assumption, it follows from~\citet[Proposition IV.9]{Cherukuri:2019} that
\begin{align} 
\label{eq:cvar-uniform-bound}
&\Pr\bigg[ \sup_{f \in \FCal} | \mathrm{CVaR}( \mathsf{L}_f ) - \mathrm{CVaR}( \hat{\mathsf{L}}_f ) | > \epsilon \bigg] \leq \gamma( \epsilon ) \cdot e^{-N \cdot \beta( \epsilon )},
\end{align} 
where $\gamma( \epsilon ) = \mathscr{O}( \mathrm{diam}( \FCal ) \cdot (\epsilon \cdot \alpha)^{-1} )$, 
and $\beta( \epsilon ) = \mathscr{O}( \alpha \cdot \epsilon^2 \cdot B^{-2} )$,
where $\mathrm{diam}( \FCal )$ is the diameter of the function class,
and $B = \|\ell\|_{\infty}$ is the maximum possible CVaR value.
Now, for a large enough constant $\mu$, by setting
\begin{align}
\label{eq:step2}
\epsilon = M \defEq \mathscr{O}\left(\min\bigg\{\frac{\mathrm{diam}(\FCal)}{\alpha\delta}, B\cdot \sqrt{\frac{1}{\alpha N}\cdot\log\bigg(\frac{\mathrm{diam}(\FCal)}{\alpha \delta}\bigg)}\bigg\}\right)
\end{align}
we obtain that
\begin{align}
\label{eq:step3}
\Pr\bigg[ \sup_{f \in \FCal} | \mathrm{CVaR}( \mathsf{L}_f ) - \mathrm{CVaR}( \hat{\mathsf{L}}_f ) | > \epsilon \bigg] \leq \delta.
\end{align}
Thus, it follows from \eqref{eq:step0}, \eqref{eq:step2}, and \eqref{eq:step3} that with probability at least $1- \delta$,
\begin{align}
\label{eq:step5}
\mathrm{CVaR}_{\alpha}( \mathsf{L}_f ) \leq \mathrm{CVaR}_{\alpha}( \hat{\mathsf{L}}_f ) \; + \OCal\left(\min\bigg\{\frac{\mathrm{diam}(\FCal)}{\alpha\delta}, B\cdot\sqrt{\frac{1}{\alpha N}\cdot\log\bigg(\frac{\mathrm{diam}(\FCal)}{\alpha \delta}\bigg)}\bigg\}\right).
\end{align}

\textbf{Rademacher bound}.
Define
$$ L_{\alpha}( f, t; \DCal ) \defEq \E{}{ \frac{1}{\alpha} [ \ell( \Y, f( \X ) ) - t ]_+ + t }. $$
We thus have 
$\mathrm{CVaR}_{\alpha}( \mathsf{L}_f ) = L_{\alpha}( f, t^*; \DCal )$ 
where $t^*$ is the $(1 - \alpha)$th quantile of the loss values.
We similarly have the empirical version
$$ \hat{L}_{\alpha}( f, t; S ) \defEq \frac{1}{N \alpha} \sum_{n = 1}^N [ \ell( \Y, f( \X ) ) - t ]_+ + t, $$
where again
$\mathrm{CVaR}_{\alpha}( \widehat{\mathsf{L}}_f ) = L_{\alpha}( f, \hat{t}^*; S )$,
for $\hat{t}^*$ the empirical quantile of the loss values.

For $\bar{\ell}_{\alpha, t}( y, f ) \defEq \frac{1}{\alpha} [ \ell( y, f ) - t ]_+ + t$,
we may write
$$ L_{\alpha}( f, t; \DCal ) = \E{}{ \bar{\ell}_{\alpha, t}( \Y, f( \X ) ) }, $$
and similarly for the empirical version.
Consequently, a standard analysis yields that with probability at least $1 - \delta$,
$$ L_{\alpha}( f, t; \DCal ) \leq \hat{L}_{\alpha}( f, t; S ) + 2 \cdot \mathrm{Rad}_N( \bar{\FCal}_{t} ) + \sqrt{\frac{\log \frac{1}{\delta}}{2N}}, $$
where 
$\bar{\FCal}_{\alpha} \defEq \{ ( x, y ) \mapsto \bar{\ell}_{\alpha, t}( y, f( x ) ) \colon f \in \FCal, t \in [ 0, B ] \}$.

The function $z \mapsto \frac{1}{\alpha} \cdot [ z ]_+$ is clearly $\frac{1}{\alpha}$-Lipschitz,
and so by the contraction principle,
$\mathrm{Rad}_N( \bar{\FCal}_{\alpha} ) \leq \frac{1}{\alpha} \cdot \mathrm{Rad}_N( \tilde{\FCal} )$,
where
$\tilde{\FCal} \defEq \{ ( x, y ) \mapsto \ell( y, f( x ) ) - t \colon f \in \FCal, t \in [ 0, B ] \}$.
Following~\citet[Theorem 1]{Boyd:2012},
we have that $\mathrm{Rad}_N( \tilde{\FCal} ) \leq \mathrm{Rad}_N( {\FCal}_{\ell} ) + \frac{B}{\sqrt{N}}$,
for 
${\FCal}_{\ell} \defEq \{ ( x, y ) \mapsto \ell( y, f( x ) ) \colon f \in \FCal \}$.
Applying the contraction principle once more,
by the assumption that $\ell$ is 1-Lipschitz,
$\mathrm{Rad}_N( {\FCal}_{\ell} ) \leq \mathrm{Rad}_N( {\FCal} )$.
It thus follows that for every $f \in \FCal$ and $t \in [ 0, B ]$,
$$ L_{\alpha}( f, t; \DCal ) \leq \hat{L}_{\alpha}( f, t; S ) + \frac{1}{\alpha} \cdot \left( \mathrm{Rad}_N( \FCal ) + \frac{B}{\sqrt{N}} \right) + \sqrt{\frac{\log \frac{1}{\delta}}{2N}}. $$

Since the above holds uniformly for every $t \in [ 0, B ]$,
we may plug in $t$ to be the population-level loss quantile.
For the right hand side, one may bound this (following Lemma~\ref{lemm:cvar-bias})
by the choice of the empirical quantile,
plus the deviation between the empirical and true quantile.
Thus, we arrive at
$$ \mathrm{CVaR}_{\alpha}( \mathsf{L}_f ) \leq \mathrm{CVaR}_{\alpha}( \hat{\mathsf{L}}_f ) \; + \frac{1}{\alpha} \cdot \left( \mathrm{Rad}_N( \FCal ) + \frac{B}{\sqrt{N}} \right) + \sqrt{\frac{\log \frac{1}{\delta}}{2N}} + \OCal\left( \frac{1}{\sqrt{N}} \right). $$
\end{proof}